# Simulating Ankle Torque during Walking Using a new Bioinspired Muscle Model with Application for Controlling a Powered Exoskeleton

[1] Safoura Sadegh Pour Aji Bishe, [1]Dan Rivera1, [2]Katherine Strausser, [1]Zachary Lerner, and [1]Kiisa Nishikawa

[1]Northern Arizona University, Flagstaff, AZ, USA
[2]Ekso Bionics Company, Richmond, CA, USA
Email: kiisa.Nishikawa@nau.edu

## INTRODUCTION

Human-like motion is a primary goal for many robotic assistive devices. Emulating the strategy of the human neuromuscular system may aid the control of such powered devices, yet many challenges remain.

Among all muscle models, Hill type models are the most commonly used for controlling assistive devices [1]. The Hill muscle model represents muscle function through two simple elements, and often fails to accurately predict muscle force in different situations [2]. Nishikawa et al. developed a novel "winding filament" hypothesis for muscle contraction (Fig. 1-A) that incorporates a role for the giant titin protein in active muscle [3].

In this study, we investigated the potential for using the winding filament model (WFM) of muscle to predict the net muscle moment of the ankle. The long-term goal is to use this model to improve ankle control of a commercial powered exoskeleton.

## METHODS

As biological tissues, muscle and tendon exhibit time dependent properties. However, Hill-type muscle models do not contain any time-dependent properties, suggesting that Hill models cannot accurately predict muscle function when time-dependent properties of muscle plays a significant role in motion. In contrast, the winding filament hypothesis incorporates time-dependent tissue behavior and may be more capable of capturing human-like actuation [3].

The WFM has a series-elastic passive component (tendon), represented by a spring in the mechanical model ($K_{ss}$ spring in Fig. 2, similar to $l^T$ in Hill model), in series with a pulley representing actin thin filaments. The contractile element is parallel to a damper ($C_{ce}$) that accounts for time-dependent properties of muscle. These two components, connected to the titin elastic element, also represented by a spring ($K_{ts}$ spring in Figure 1-A).

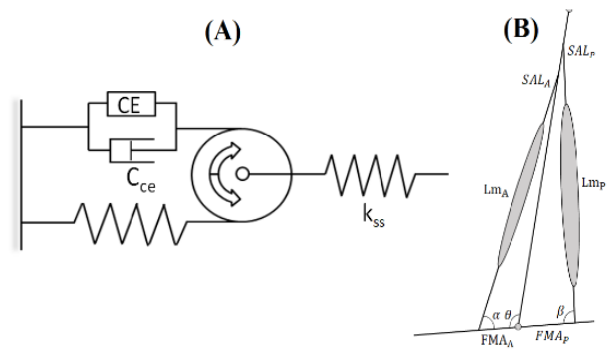

*Figure 1*: (*A*) *Mechanical model of winding filament hypothesis, that contains two springs for representing the series elastic component ($K_{ss}$) and titin elastic component ($K_{ts}$). There is also contractile element ($CE$) parallel to a damper ($C_{ce}$).* (*B*) *Two virtual muscles used in model of ankle joint. One represents the anterior muscles ($m_A$), and the other one represents the posterior muscles ($m_P$). ($Lm_A$) and ($Lm_P$) are the length of virtual anterior muscle and virtual posterior muscle, respectively, calculated from the muscle attachments and ankle joint angle (θ).*

We use lump-sum model design, where only two muscles were used, one representing all the anterior muscles that contribute to ankle dorsiflexion, and the other one represents all the posterior muscles (Figure 1-B) that contribute to plantarflexion.

We evaluated the ability of our model to accurately predict the ankle moment by adjusting only a single activation node of the muscle activation signal by comparing model output to the ankle moment computed from Inverse Dynamics (ID) in OpenSim.

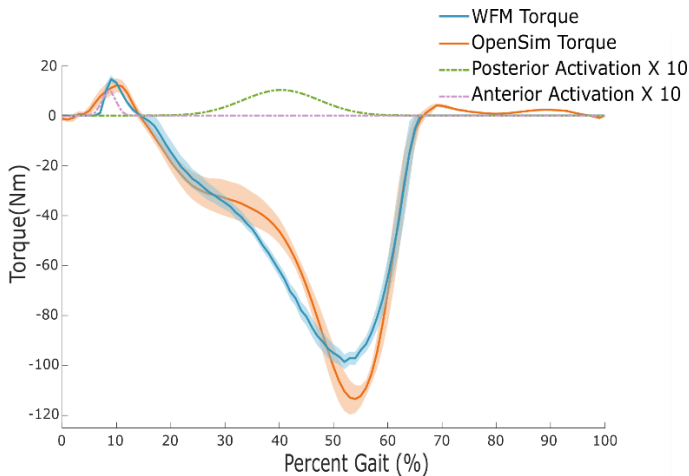

*Figure 2*: The result of optimization for one male subject. The blue line is the predicted torque by WFM, and the red line is the torque estimated by OpenSim. The dash point graphs shows the activation for anterior and posterior virtual muscle.

We used Particle Swarm Optimization (PSO) to minimize the root mean square error (RMSE) between the ankle moment from Inverse Dynamics and the output torque of the model. The optimization where done using the walking data of four healthy subjects (two male and two female) as training data, and then we test the results of optimization on another four subjects (one male and three female). The walking speed of all the patients were 125 (m/s).

**RESULTS AND DISCUSSION**

The control of exoskeleton should be designed generally for all populations, so the activation of muscles should be defined generally too. In this case, we designed a general shape for activation that is a function of gait cycle time. The result of the optimization shows that the best activation curve that fits for the training walking data of four subjects has the maximum activation of 0.05 for anterior virtual muscle, and 0.1 for posterior one. The shape of the resulting activations is also shown in Figure 2 by dash lines. Based on the result of optimization, the maximum force of the muscle was calculated 5 times of subject's body weight in kilogram. The RMSE between WFM torque and Inverse Dynamics in OpenSim for test and train data are presented in Table 1.

The purpose of this study was evaluating two general activation curves that best fit the torque output of WFM model to the ID in OpenSim. It is expected that the RMSE of the WFM modeling be large, since it is a general model that should work for every different person with different pattern of walking. The important result of this study is that the test data have the same amount of RSME as the train data—the mean of RSME for train subject is 154.78, and for test subject 129.07. Comparing the RMSE of test and train data indicates that our model can predict the ankle joint torque of all populations with the same amount of error, which is acceptable for an exoskeleton.

*Table 1:* The root mean square error (RMSE) between the torques calculated from WFM and ID in OpenSim.

| Type of data | Train | Train | Train | Train | Test | Test | Test | Test |
|---|---|---|---|---|---|---|---|---|
| Subject number | 1-F* | 2-F | 3-M* | 4-M | 1-F | 2-F | 3-M | 4-F |
| RMSE | 118.95 | 167.23 | 213.85 | 119.08 | 102.16 | 110.88 | 163.59 | 139.67 |

*F stands for Female subject, and M stands for Male subject.